\documentclass[10pt,twocolumn,letterpaper]{article}

\usepackage[pagenumbers]{cvpr} % To force page numbers, e.g. for an arXiv version
\usepackage{graphicx}
\usepackage{amsmath}
\usepackage{amssymb}
\usepackage{booktabs}
\usepackage{color}
\usepackage{colortbl}  
\usepackage{xcolor}
\usepackage{array}
\usepackage{bbding}
\usepackage{subfiles}

\definecolor{hollywoodcerise}{rgb}{0.96, 0.0, 0.63}
\definecolor{lasallegreen}{rgb}{0.03, 0.47, 0.19}
\definecolor{hanpurple}{rgb}{0.32, 0.09, 0.98}
\definecolor{green(pigment)}{rgb}{0.0, 0.65, 0.31}
\usepackage{multirow}
\definecolor{cvprblue}{rgb}{0.21,0.49,0.74}
\usepackage[pagebackref,breaklinks,colorlinks,citecolor=cvprblue]{hyperref}

\def\paperID{6652} %{6652} % *** Enter the Paper ID here
\def\confName{CVPR}
\def\confYear{2024}

\title{EIT-1M: One Million EEG-Image-Text Pairs for Human Visual-textual Recognition and More}

\author{Xu Zheng$^{1}$ \thanks{$^\dagger$ equal contribution. $^\ddagger$ corresponding author.} \quad Ling Wang$^{1}$ \footnotemark[1] \quad Kanghao Chen$^{1}$ \footnotemark[2] \quad Yuanhuiyi Lyu$^{1}$ \footnotemark[2] \quad Jiazhou Zhou$^{1}$ \quad Lin Wang$^{1}$$^{,2}$ \footnotemark[3]% \thanks{Corresponding author.}
\\
$^{1}$AI Thrust, HKUST(GZ) \quad $^{2}$Dept. of CSE, HKUST 
\\
{\tt\small \{yuanhuiyilv, jiazhouzhou\}@hkust-gz.edu.cn, zhengxu128@gmail.com, linwang@ust.hk}
}

\begin{document}
\maketitle

\begin{abstract}
Recently, electroencephalography (EEG) signals have been actively incorporated to decode brain activity to visual or textual stimuli and achieve object recognition in multi-modal AI. Accordingly, endeavors have been focused on building EEG-based datasets from visual or textual single-modal stimuli. However, these datasets offer limited EEG epochs per category, and the complex semantics of stimuli presented to participants compromise their quality and fidelity in capturing precise brain activity.  The study in neuroscience unveils that the relationship between visual and textual stimulus in EEG recordings provides valuable insights into the brain's ability to process and integrate multi-modal information simultaneously. Inspired by this, we propose a novel large-scale multi-modal dataset, named \textbf{EIT-1M}, with over 1 million EEG-image-text pairs. Our dataset is superior in its capacity of reflecting brain activities in simultaneously processing multi-modal information. To achieve this, we collected data pairs while participants viewed alternating sequences of visual-textual stimuli from 60K natural images and category-specific texts. Common semantic categories are also included to elicit better reactions from participants' brains. Meanwhile, response-based stimulus timing and repetition across blocks and sessions are included to ensure data diversity. To verify the effectiveness of EIT-1M, we provide an in-depth analysis of EEG data captured from multi-modal stimuli across different categories and participants, along with data quality scores for transparency. We demonstrate its validity on two tasks: 1) EEG recognition from visual or textual stimuli or both and 2) EEG-to-visual generation. 
\end{abstract}

\section{Introduction}
%\vspace{-5pt}

Electroencephalography (EEG) is a widely applied neuroimaging modality in cognitive neuroscience. It is known for its ability to decipher intricate brain activity patterns during various cognitive processes~\cite{saeidi2021neural}. In the early days, research focused on constructing EEG datasets for medical purposes, such as detecting and predicting seizures~\cite{wong2023eeg}. 
Recently, EEG signals have been broadly incorporated to decode brain activity to visual or textual stimuli and achieve object recognition in multi-modal artificial intelligence (AI)~\cite{teplan2002fundamentals, cohen2017does, benchetrit2023brain, roy2019deep, singh2023eeg2image, singh2024learning}. 
This enriches the data landscape, allowing for more nuanced and accurate models of brain activity and cognitive processes. 
% This integration is pivotal for advancing both bio-inspired AI technologies and our understanding of the human brain.
% Recent advancements in artificial intelligence (AI) have been propelled by large multi-modal foundation models~\cite{girdhar2023imagebind, lyu2024unibind}, which aim to harness the potential of diverse modalities in open-world environments. For example, ViT-Lens~\cite{lei2023vit} integrates brain activity signals, specifically Electroencephalography (EEG), a widely used neuroimaging modality in cognitive neuroscience and medical imaging, to decode brain activity and meanwhile achieve object recognition~\cite{teplan2002fundamentals, cohen2017does, benchetrit2023brain, roy2019deep, singh2023eeg2image, singh2024learning}. 
% EEG is a widely used neuroimaging modality in cognitive neuroscience and medical imaging, known for its ability to decipher intricate brain activity patterns during various cognitive processes~\cite{teplan2002fundamentals, cohen2017does, benchetrit2023brain, roy2019deep, singh2023eeg2image, singh2024learning}.
% 

% However, replicating the success of data-driven multi-modal AI algorithms for rarer modalities is challenging due to data scarcity, particularly the lack of comprehensive paired datasets involving images, text, and EEG signals. 
Accordingly, research endeavors have been focused on building EEG-based datasets
% from either visual or textual single-modal stimuli
~\cite{hollenstein2019zuco,gifford2022large, kavasidis2017brain2image, tirupattur2018thoughtviz, THINGSEEG1}, as summarized in Tab.~\ref{tab:related}.
% Additionally, there exist EEG-Text datasets~\cite{hollenstein2019zuco} and EEG-Image datasets~\cite{gifford2022large, kavasidis2017brain2image, tirupattur2018thoughtviz, THINGSEEG1, grootswagers2021things} that capture EEG waveforms recorded while participants are exposed to textual or visual stimuli. 
For instance, ZuCo 1.0~\cite{hollenstein2018zuco} is a pioneering EEG-Text dataset that records neural processes underlying reading and language comprehension during the reading tasks. On the other hand, 
Brain2Image~\cite{KavasidisPSGS17} is a representative EEG-image dataset that includes evoked responses to visual stimuli from 40 classes. 
However, these datasets have two distinct shortcomings: \textbf{\textit{1)}} They offer limited EEG epochs per category, and the complex semantics of stimuli presented to participants compromise their quality and fidelity in capturing precise brain activity. \textbf{\textit{2}) }They only encompass EEG signals recorded from single-modal stimuli, either visual or textual. This makes them less possible to be used for training high-performance multi-modal AI models.
% from scratch. 
% To address these limitations, a more integrated approach encompassing EEG, images, and text is essential for advancing AI and neuroscience research.
\begin{table*}[t]
% sensor / modality / 
    \centering
    \renewcommand{\tabcolsep}{4pt}
    \renewcommand\arraystretch{1.6}

    \resizebox{\linewidth}{!}{
    \begin{tabular}{l|c|c|c|c|c|c|c}
    \hline
    Dataset  & Year & Equipment & Modality & Epochs & CEA & MEA & Purpose \\ \hline % & Stimuli 
    Brain2Image~\cite{kavasidis2017brain2image} & 2017 & Brainvision BrainAmp DC & EEG-Image  & 11,466 &$\times$&$\times$ & Decoding \\ \hline %  & ImageNet~\cite{deng2009imagenet}
    EVSR~\cite{kumar2018envisioned} & 2018 & Emotiv EPOC+ & EEG-Image  & 13,800 &\checkmark&$\times$ &Recognition\\ \hline %  & ImageNet~\cite{deng2009imagenet}
    ZuCo 1.0~\cite{hollenstein2018zuco} & 2018 & EGI Geodesic Hydrocel system & EEG-Text  & 259,788 &$\times$&$\times$ &NLP\\ \hline % &Recogonition & ImageNet~\cite{deng2009imagenet}
    ZuCo 2.0~\cite{hollenstein2019zuco} & 2020 & EGI Geodesic Hydrocel system& EEG-Text  & 272,484 &$\times$&$\times$ &NLP\\ \hline % &Recogonition &
    THINGS EEG1~\cite{THINGSEEG1}& 2022 & Brainvision actiCHamp & EEG-Image &1,112,400 &\checkmark&$\times$ &Recognition\\ \hline % &Recogonition & THINGS~\cite{hebart2019things} 
    THINGS EEG2~\cite{THINGSEEG2}& 2022 & Brainvision actiCHamp & EEG-Image  &821,600 &$\times$&$\times$ &Recognition\\ \hline % &Recogonition & THINGS~\cite{hebart2019things}
    Alljoined1~\cite{alljoined1} & 2024 & BioSemi ActiveTwo & EEG-Image  & 46,080 &$\times$&$\times$ &Decoding\\ \hline % &Decoding & MS-COCO~\cite{lin2014microsoft}
    \textbf{EIT-1M (Ours)} & 2024 & Brainvision actiCHamp Plus & EEG-Image-Text &1,200,000 &\checkmark&\checkmark &Recognition \& Decoding \\ % &Recogonition & CIFAR-10~\cite{krizhevsky2014cifar} 
    \hline
    \end{tabular}}
    \caption{EEG datasets. (CEA: Category-level ERP Analysis, MEA: Multi-modal ERP Analysis.)}
    \label{tab:related}
        %\vspace{-12pt}
\end{table*}

The study in neural science reveals that EEG recordings reveal a significant relationship between visual and textual stimuli, offering valuable insights into the brain's capacity to integrate multi-modal information simultaneously~\cite{saeidi2021neural}. This integration is crucial for understanding how the brain processes complex, real-world scenarios where multiple types of sensory input are encountered simultaneously. 
Inspired by this, we introduce a novel large-scale multi-modal EEG dataset, \textbf{EIT-1M}, comprising paired EEG, visual, and textual data for the benefit of research communities.
The key insight of our dataset is to record human brain activities while simultaneously processing multi-modal information. To achieve this, data was collected from five participants exposed to random sequences of 60K natural images and their corresponding category descriptions. To date, we have gathered over \textbf{1 million} epochs of brain responses using a 64-channel EEG headset (actiCHamp Plus\footnote{\url{https://brainvision.com/products/actichamp-plus/}}).

\begin{figure*}[h!]
    \centering
    \includegraphics[width=\textwidth]{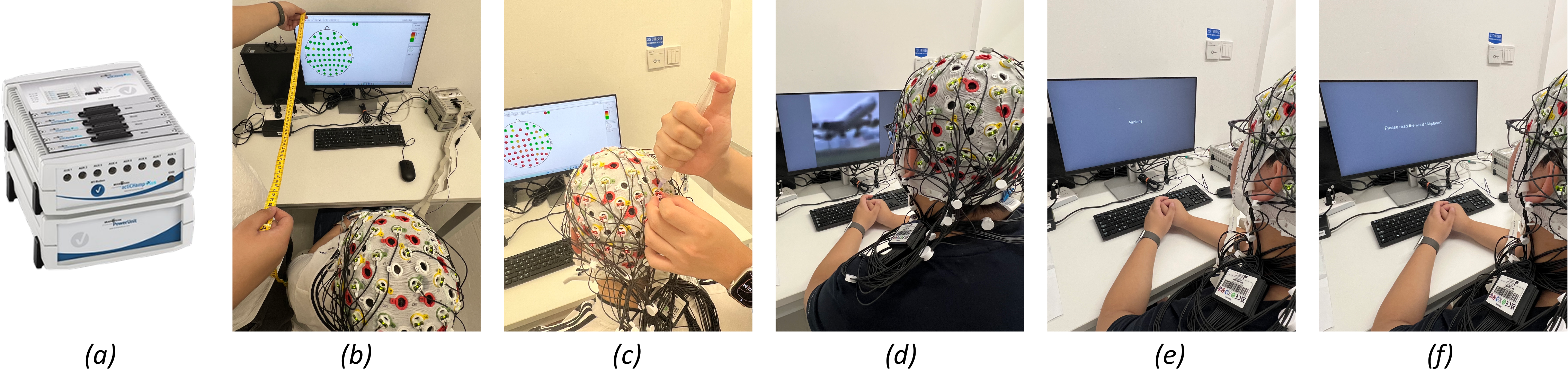}
    %\vspace{-20pt}
    \caption{(a) actiCHamp Plus device. (b) Experimental setup with monitor 80 cm from participant. (c) Injecting conductive gel. (d) Visual stimuli. (e) Textual stimuli. (f) Speech stimuli.}
    \label{fig:procedure}
    %\vspace{-6pt}
\end{figure*}

\begin{figure*}[t!]
    \centering
    \includegraphics[width=\textwidth]{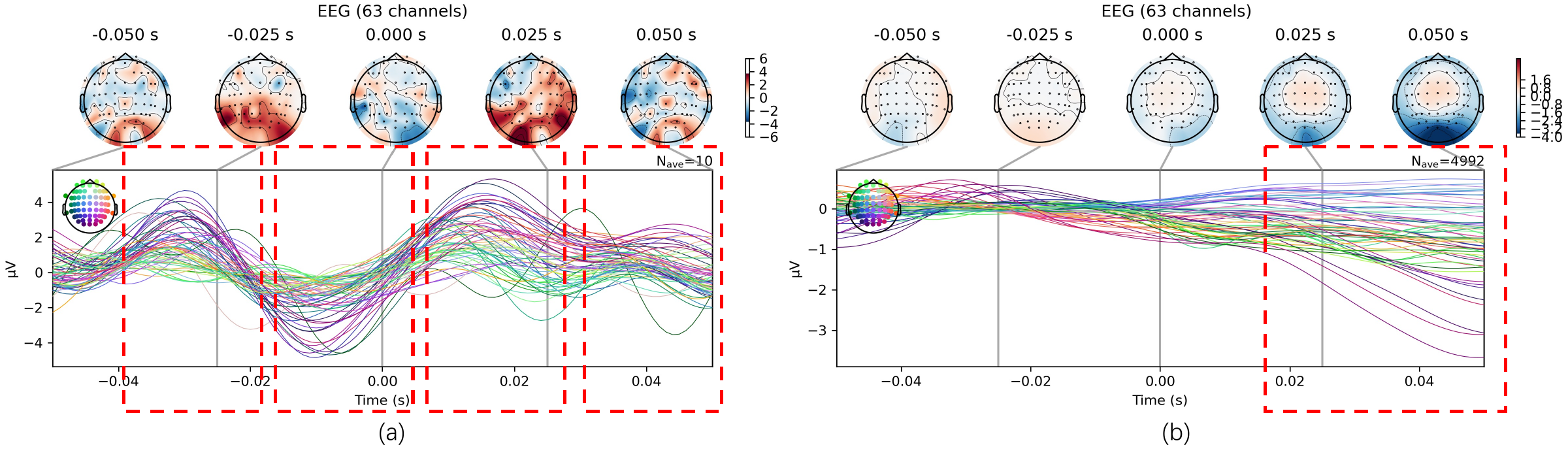}
    %\vspace{-20pt}
    \caption{(a) EEG signals from high-resolution visual stimuli.% exhibit more complex and higher amplitude fluctuations, indicating increased processing time and effort. 
    (b) EEG signals from our visual stimuli.%, with a moderate resolution, demonstrate balanced neural responses, clear ERP components, and consistent patterns. 
    }
    \label{fig:resolution}
        %\vspace{-10pt}
\end{figure*}

Specifically, we utilize the 10-category dataset CIFAR-10~\cite{krizhevsky2009learning} to construct the visual and textual stimulus. This dataset harnesses an image resolution of 32×32 pixels without excessive details,
Empirically, as shown in Fig.~\ref{fig:resolution}, we find that low-resolution visual stimuli stimulate more stable neural responses, suggesting they are appropriate and manageable within a brief viewing period. We present visual and textual stimuli sequentially to maintain continuous engagement with the objects and concepts, as shown in Fig.~\ref{fig:rsvp_total}.
% It is important to note that all EEG signals for a single category, including both image and text stimuli as shown in Fig.~\ref{fig:stimuli}, were captured within the same block, maintaining the quality and paired nature of our multi-modal dataset. 
Moreover, our dataset features response-based stimulus timing, repetition across blocks and sessions, and diverse visual and textual classes. 
% As such, we aim to provide a comprehensive  
To verify the effectiveness, we provide an in-depth analysis of EEG data captured from multi-modal stimuli across different categories and participants. The data analysis includes EEG topographic maps, corresponding signals analysis and ERP analysis. These analysis highlight the distinct ERP characteristics from visual and textual stimuli, providing insights in the multi-modal information processing of brains.
For transparency, we include data quality scores (See Tab.~\ref{tab:perclasssnr}).

To benchmark our EIT-1M, we demonstrate its validity on two tasks: 1) EEG recognition from visual or textual stimuli or both (See Sec.~\ref{sec:recogntion}) and 2) EEG-to-visual generation (See Sec.~\ref{sec:generation}). 
% We have released part of the dataset and code at \url{https://eit-1m.github.io/EIT-1M/} for anonymous review, with all data to be publicly available upon acceptance. 
We expect our dataset to be a benchmark contributor for advancing the research for multi-modal AI~\cite{ lyu2024omnibind,zheng2024eventdance,zhou2024exact,lyu2024image,cao2023chasing,chen2023clip,zhou2023clip,zheng2023deep} and potentially for cognitive neuroscience. 
% The contributions of our paper can be summarized as follows:(I) 

% Brain activities can be recorded using techniques such as EEG, which possess spatial and temporal resolutions that enable the decoding of specific visual and linguistic stimuli~\cite{HuthHGTG16, SpampinatoPKGSS17}.

%\vspace{-10pt}
\section{Related Work}
\noindent \textbf{EEG Datasets with Visual Stimuli.}
They capture EEG waveforms while participants view visual stimuli, facilitating studies of brain activity, as shown in Tab.~\ref{tab:related}. A representative dataset is Brain2Image~\cite{KavasidisPSGS17}, which includes evoked responses to visual stimuli from 40 classes, each with 50 images, totaling 2K images. However, this dataset is impeded by its lack of train-test separation during recording, block-specific stimuli patterns, and inconsistency across frequency bands~\cite{BharadwajWS23, LiJAIWBS21}. In contrast, the THINGS-EEG1~\cite{THINGSEEG1} and THINGS-EEG2~\cite{THINGSEEG2} datasets address these issues by incorporating both main and validation sessions to ensure data quality and consistency. These two datasets contain human EEG responses from 50 subjects to 22,248 images in the THINGS stimulus set.

Regarding the diversity of stimuli, while studies like Brain2Image and ~\cite{AhmedWBS21} involve 40 classes, other studies focus on only 10 different image classes~\cite{tirupattur2018thoughtviz}. This limited representation allows for more controlled studies but fails to capture the continuous and diverse nature of naturalistic stimuli due to the limited samples from each category. Other datasets like MindBigData~\cite{mindbigdata} and ~\cite{AhmedWBS21} capture a wide range of images but are derived from a single individual, limiting their potential for training image reconstruction models that generalize to other individuals. Recently, to address these limitations, Alljoined1~\cite{alljoined1} includes 10K images per participant from object categories in MS-COCO~\cite{LinMBHPRDZ14}, thereby accounting for the diversity and continuity of real-world images.

\noindent \textbf{EEG Datasets with Textual Stimuli} 
They are primarily developed for brain signal decoding. Notable examples include ZuCo 1.0~\cite{hollenstein2018zuco} and ZuCo 2.0~\cite{hollenstein2019zuco}, captured with 128-channel EEG devices. These datasets provide insights into the neural processes underlying reading and language comprehension by recording EEG signals during reading tasks. EEG2Text~\cite{DBLP:conf/icml/LiuZDZP19} focuses on translating brain signals into textual descriptions, supporting the development of AI models for decoding and generating text from EEG signals.
Despite these advancements, there remains a need for datasets that integrate both visual and textual stimuli to capture the complex interplay between different modalities in the brain. In a nutshell, all these datasets primarily focus on single-modal stimuli, limiting their fidelity for training multi-modal AI models. \textit{Our EIT-1M dataset addresses this gap by providing paired EEG, visual, and textual data, enabling comprehensive multi-modal analysis. Thus our dataset is superior in its capacity of reflecting brain activities in simultaneously processing multi-modal information.}
% and fostering advancements in both AI and cognitive neuroscience.

\section{Dataset Collection Methods}
% The overview of our dataset is shown in 
Tab.~\ref{tab:dataset_overview} provides an overview of one experiment involving five participants, aged 20-30 years, with a gender distribution of two females and four males. Each participant underwent two 300-minute sessions, during which 1,200,000 events were recorded, including 600K visual and 600K textual stimuli. The stimuli were drawn from ten CIFAR-10 categories for visuals and ten textual categories. EEG recordings were made using a 64-channel headset at a 1000 Hz sampling rate. The dataset ensures high quality with an average signal-to-noise ratio as in Tab.~\ref{tab:perclasssnr}, maintaining impedance levels at or below 20 k$\Omega$. Each session featured an average of 10K events, with each event lasting 50 ms and an inter-event interval of 1 second. Preprocessing involved 1-40 Hz band-pass filtering and epoching from -20 to 30 ms relative to stimulus onset, with baseline correction at -20 ms. This dataset aims to support research in EEG analysis and multi-modal recognition.

%\vspace{-6pt}
\subsection{Experimental Settings}
%\vspace{-6pt}
\noindent\textbf{Participants}
Five adults (mean age 24.83 years; 1 female, 4 male) participated in this study, all with normal or corrected-to-normal vision, and none of them have suffered or are suffering neurological or psychiatric problems such as ADHD and epilepsy. Each participant provided informed written consent and received monetary reimbursement for their involvement. 
% The study procedures were approved by the ethical committee of the Information Hub at the Hong Kong University of Science and Technology (Guangzhou).
The study procedures were approved by the ethical committee.
It is important to acknowledge the potential limitations of this study, such as the gender imbalance among participants and the low age disparity. 

% These factors could introduce bias and affect learning outcomes in AI models.

\begin{table*}[t!]
\centering
\footnotesize
    \renewcommand{\tabcolsep}{6pt}
\resizebox{\linewidth}{!}{
\begin{tabular}{>{\raggedright\arraybackslash}p{0.25\linewidth} >{\raggedright\arraybackslash}p{0.25\linewidth} >{\raggedright\arraybackslash}p{0.35\linewidth}}
\toprule
\textbf{Item} & \textbf{Description} & \textbf{Details} \\ 
\midrule
\textbf{Participants} & Number & 5 \\ 
 & Age Range & 20-30 years \\ 
 & Gender Distribution & 1 females, 4 males \\ 
\midrule
\textbf{Sessions} & Number per Participant & 2 \\ 
 & Duration & 4 hours each \\ 
\midrule
\textbf{Total Events} & Total & 1,200,000 \\ 
 & Visual Stimuli & 600,000 \\ 
 & Text Stimuli & 600,000 \\ 
%  & Speech Stimuli &  \\ 
\midrule
\textbf{Stimuli} & Categories & 10 from CIFAR-10 \\ 
 & Description & Visual: Images from CIFAR-10; Text: category names \\ 
\midrule
\textbf{Recording Details} & Sampling Rate & 1000 Hz \\ 
 & EEG Channels & 64 \\ 
 & Equipment & actiCHamp Plus\footnote{\url{https://brainvision.com/products/actichamp-plus/}} \\ 
\midrule
\textbf{Data Quality} 
% & Average SNR & 30 dB \\ 
%  & Usable Data & 95\% \\ 
 & Impedance Levels & $\leq$ 20 k$\Omega$ \\ 
\midrule
\textbf{Event Details} & Average Events/Session & 120,000 \\ 
 & Event Duration & 50 ms \\ 
 & Inter-event Interval & 50 ms \\ 
\midrule
\textbf{Preprocessing} & Filtering & 1-40 Hz band-pass \\ 
 & Epoching & -50 to 50 ms relative to stimulus onset, baseline correction at -50 ms \\ 
\bottomrule
\end{tabular}
}
\caption{Overview of our proposed EEG-Image-Text Dataset}
  %\vspace{-5pt}
%\vspace{-8pt}
\label{tab:dataset_overview}
\end{table*}

\begin{table*}[t!]
    \centering
  
      %\vspace{-5pt}
    \renewcommand{\tabcolsep}{12pt}
    \resizebox{\linewidth}{!}{
    \begin{tabular}{ccccccccccc}
    \toprule
    Category & Airplane & Automobile & Bird & Cat & Deer & Dog & Frog & Horse & Ship & Truck \\ \midrule
    Average SNR / dB (raw data) & 6.14 & 5.76 & 6.09 & 5.69 & 5.44 & 4.82 & 4.47 & 4.74 & 4.30 & 3.67 \\
    % Usable Data & \\ 
    \bottomrule
    \end{tabular}}
    \caption{Example raw data quality of participant 04 in our proposed dataset across different blocks (categories) within the first session.}
    \label{tab:perclasssnr}
        %\vspace{-16pt}
\end{table*}

\noindent\textbf{Stimuli.}
All images used in this study as visual stimuli are sourced from the CIFAR-10 dataset~\cite{krizhevsky2009learning}. This dataset is a well-known benchmark in machine learning and computer vision, comprising 60K color images across 10 different classes, with 6K images per class.
These classes represent a variety of everyday objects and animals, including airplanes, automobiles, birds, cats, deer, dogs, frogs, horses, ships, and trucks.
This dataset is selected for our study due to its diversity and balanced categories, which provide a robust set of stimuli for examining neural responses across different visual contexts. Utilizing this dataset allows for an in-depth exploration of how the brain processes various types of visual information and supports the development of multi-modal models that can generalize across different categories of visual stimuli. 
% Additionally, the existence of paired datasets in other modalities, such as the CIFAR-10-DVS dataset for event-based vision, further reinforces the relevance and utility of CIFAR-10 in multi-modal research.
Each image in this dataset has a resolution of 32x32 pixels, making it ideal for stimulating brain activities for visual stimuli from participants.
% testing and evaluating image classification algorithms. 
The textual stimuli are derived from the category names within the CIFAR-10 dataset: airplane, automobile, bird, cat, deer, dog, frog, horse, ship, and truck.
% The reason why we choose these images is because

\begin{figure*}[t!]
    \centering
    \includegraphics[width=\textwidth]{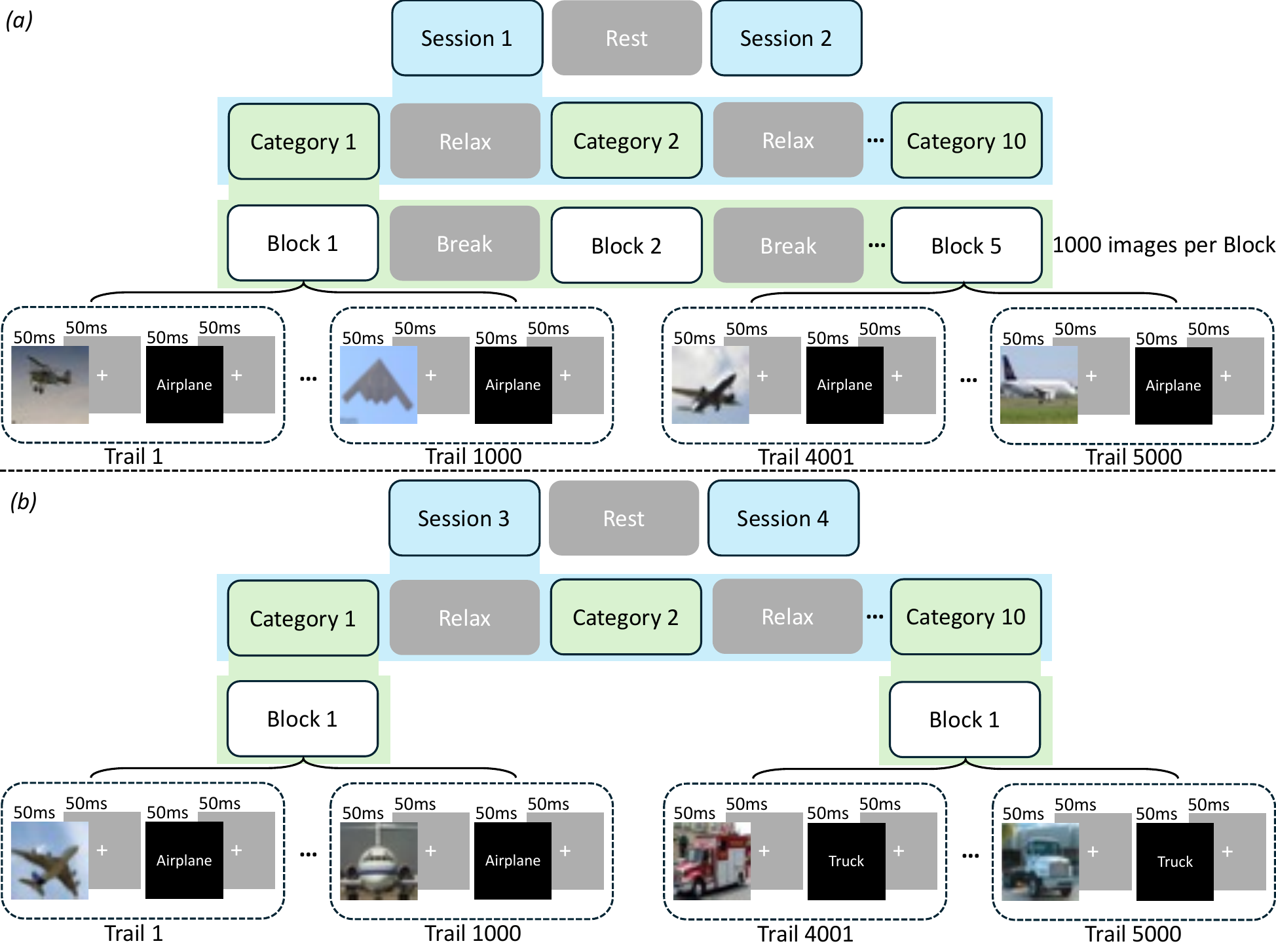}
    %\vspace{-16pt}
    \caption{Schematic overview of the structure of trials, blocks, categories and sessions with RSVP paradigm. (a) Training set of CIFAR-10 dataset; (b) Testing set of CIFAR-10 dataset.
    % (a) Each of the 5 block-specific CIFAR-10 training images and label text is presented once within each block, and each of the 10 category-specific blocks is presented once within each session. Each participant performed two sessions on different days. Each of the 2 sessions thus consists of 50,000 images and texts within and across blocks. (b) Session 3 and 4 consist of 10,000 images and labels from the CIFAR-10 testing set.
    }
    %\vspace{-20pt}
    \label{fig:rsvp_total}
\end{figure*}

% \begin{figure*}[t!]
%     \centering
%     \includegraphics[width=\textwidth]{images/rsvp_test.pdf}
%     %\vspace{-16pt}
%     \caption{Session 3 and 4 consist of 10,000 images and labels from the CIFAR-10 testing set.}
%     %\vspace{-12pt}
%     \label{fig:rsvp_test}
% \end{figure*}

\noindent \textbf{Hardware Setup}
We recorded data using a 64-electrode actiCHamp Plus system, digitized at a rate of 1024 Hz with 24-bit A/D conversion. The montage was arranged according to the international 10-20 System, and the electrode offset was kept below 40 mV. A 22-inch Dell monitor with a resolution of 1080p at 60 Hz was used to display the visual and textual stimuli. As shown in Fig.~\ref{fig:procedure}, the monitor was centrally positioned at a distance of 80 cm from the participant, maintaining a 3.5-degree angle of stimuli. We ensured that the angle remained small to minimize the occurrence of gaze drift.

%\vspace{-6pt}
\subsection{Data Collection Procedure}
%\vspace{-6pt}
Before viewing the stimuli, conductive gel was injected into each electrode to ensure the resistance was less than 20 ohms, facilitating better signal capture. Participants were then shown images and text over the course of four sessions, each four hours long. Each session comprised multiple blocks, with each block containing images from the same class and the corresponding category name text. The visual and textual stimuli were randomly arranged in a visual-textual-visual-textual order within each block. Different blocks contained stimuli from different classes. Within each block, 1,000 visual stimuli images and 1,000 text stimuli category names from CIFAR-10 were presented.

Within each trial, an image was presented for 50 ms, followed by 50 ms of a black screen. The corresponding category name of the image was also presented for 50 ms, followed by 50 ms of a black screen. A white fixation cross was visible on the screen throughout the entire trial. To ensure focus, participants were prompted to press the space bar after completing two consecutive blocks. Additionally, five to ten-minute breaks were provided between blocks based on participants' needs for better data recording.
Fig.~\ref{fig:rsvp_total} (a) shows a schematic overview of the structure of trials, blocks, categories, and sessions, which follows rapid serial visual presentation (RSVP) paradigm~\cite{THINGSEEG1,intraub1981rapid,keysers2001speed}. Each of the 5 block-specific CIFAR-10 training images and label text is presented once within each block, and each of the 10 category-specific blocks is presented once within each session. Each participant performed two sessions on different days. Each of the 2 sessions thus consists of 50,000 images and texts within and across blocks. Fig.~\ref{fig:rsvp_total} (b) illustrates that Sessions 3 and 4 consist of 10,000 images and labels from the CIFAR-10 testing set.

% \begin{figure*}[t!]
%     \centering
%     \includegraphics[width=\textwidth]{images/Topo_class.pdf}
%     \caption{EEG topographic maps for different categories. Each map represents the EEG activity at the same epoch in each trial for the respective category, highlighting distinct patterns of brain activity associated with each visual stimulus.}
%     \label{fig:topo_class}
% \end{figure*}

% %\vspace{-6pt}
\section{Data Analysis}
\begin{figure*}[t!]
    \centering
    \includegraphics[width=\textwidth]{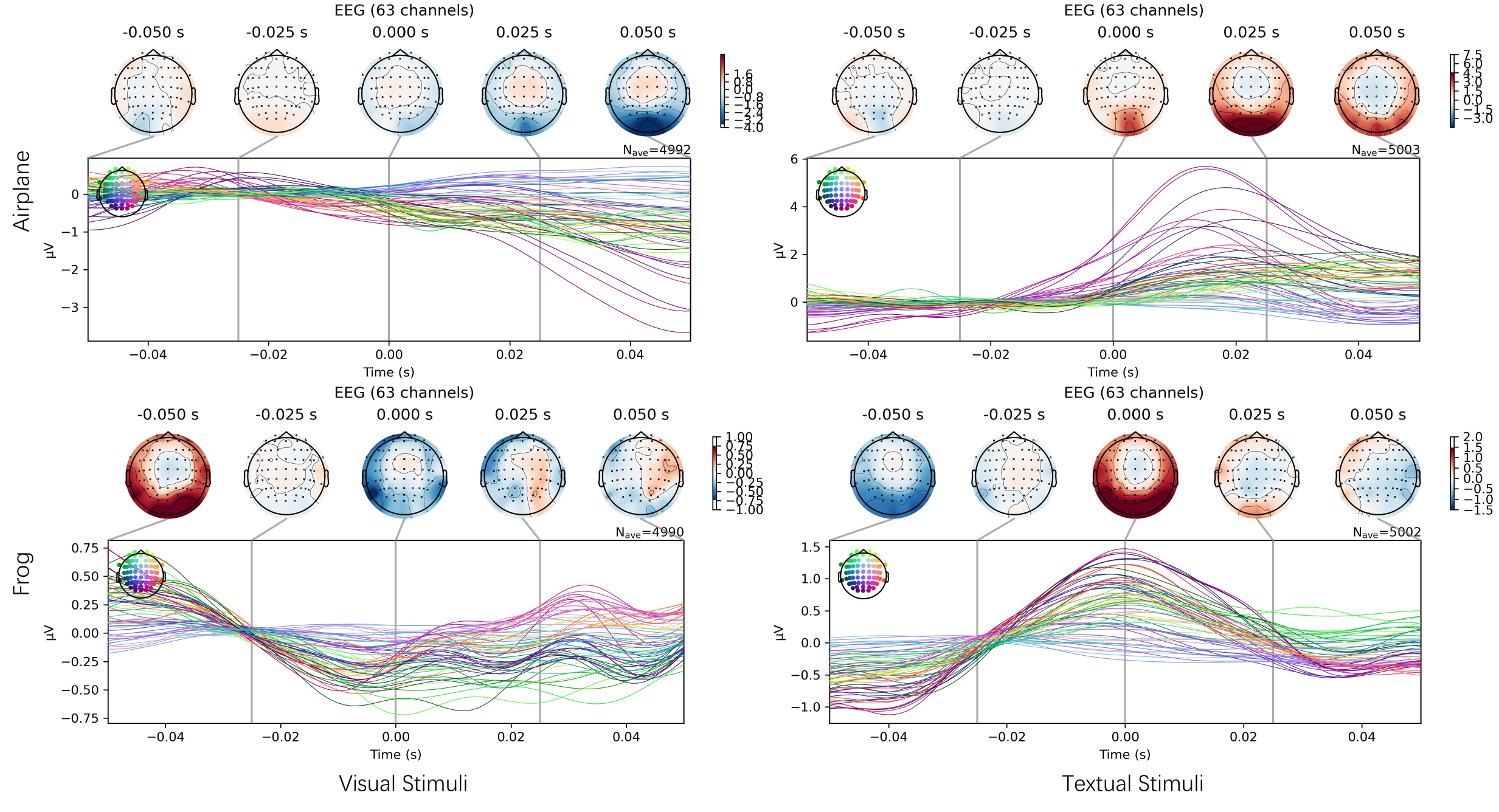}
    %\vspace{-20pt}
    \caption{EEG topographic maps and corresponding signals 
    % at all 63 electrodes (the other one as reference electrodes) 
    averaged over events for the participant viewing visual stimuli (\textbf{left column}) viewing the airplane (1st row) and frog (2nd row) images from the CIFAR-10 dataset, and events for the participant viewing textual stimuli (\textbf{right column}) viewing the airplane (1st row) and frog (2nd row) text. % from the category names of CIFAR-10 dataset. % These maps highlight individual and common brain activity patterns associated with both image and text presentation. An event is defined as a specific time point in the experiment.
    }
    %\vspace{-20pt}
    \label{fig:img_txt_compare}
\end{figure*}

\subsection{EEG Topographic Maps and Corresponding Signals Analysis}
Fig.~\ref{fig:img_txt_compare} presents the comparison of EEG signals by showcasing topographic maps and corresponding signals averaged across 63 electrodes (channel FCz as reference) for different stimuli conditions, \ie, visual and textual stimuli with airplane and frog categories. Each column of the figure represents a different stimulus type: visual stimuli (left column) and textual stimuli (right column). The visual stimuli include images from the CIFAR-10 dataset, and the textual stimuli comprise category names from the same dataset. Each row represents different categories, specifically airplane and frog. \\
\noindent \textbf{Visual Stimuli (Left Column of Fig.~\ref{fig:img_txt_compare})}
The topographic maps show the distribution of brain activity across the scalp at various time points (-0.050s, -0.025s, 0.000s, 0.025s, and 0.050s) after the stimulus onset. The maps reveal distinct patterns of neural activation, indicating how the brain processes visual stimuli over time. For instance, the airplane category (1st row) shows significant activation in the occipital and parietal regions, which are known to be involved in visual processing~\cite{robinson2017very}. The corresponding ERP signals show the average response over time for all electrodes. The signals depict the dynamic changes in brain activity, with notable peaks and troughs corresponding to different cognitive processes. For the visual stimuli, there are clear ERP components around 20ms and 40ms, which might correspond to early visual processing and higher-level cognitive processing, respectively. \\
\noindent \textbf{Textual Stimuli (Right Column of Fig.~\ref{fig:img_txt_compare})}
Similar to the visual stimuli, the topographic maps for textual stimuli show brain activity at the time points in 50 ms later as visual stimuli. Note that the visual and textual stimuli are presented with a gap of 50 ms. There are noticeable differences in the activation patterns compared to visual stimuli, highlighting the distinct neural processes involved in reading and understanding texts of the participants. For the airplane text (1st row), there is significant activation in the temporal and frontal regions, areas associated with language processing~\cite{hollenstein2021decoding}. The ERP signals for textual stimuli also display characteristic peaks, though the patterns differ from those elicited by visual stimuli. The airplane text category shows a strong response between 20 ms to 40 ms, likely reflecting early semantic processing~\cite{costanzo2013spatial}.

According to these visualizations, we have the following findings:
\textbf{(I) Individual and Common Patterns}: Fig.~\ref{fig:img_txt_compare} highlights both individual and common brain activity patterns associated with both image and text presentation. This indicates that while there are distinct neural processes for visual and textual stimuli, there are also commonalities in how the brain responds to different types of information.
\textbf{(II) Temporal Dynamics:} The temporal dynamics of the ERP signals provide insights into the timing of cognitive processes. Early components (within the last 20ms) are typically associated with sensory processing, while earlier components (before 20ms) are linked to cognitive and semantic processing.
\textbf{(III) Gap Influence}: The 50 ms gap between visual and textual stimuli presentations allowed us to observe the sequential processing of different modalities, showing how the brain transitions between visual and textual information processing.
\textbf{(IV) ERP Characteristics}: The ERP characteristics, such as the peaks around 20 ms and 40 ms for visual stimuli and between 0 ms to 20 ms for textual stimuli, provide valuable hints for understanding the stages of information processing in the brain.

\begin{figure*}[h!]
    \centering
    \includegraphics[width=\textwidth]{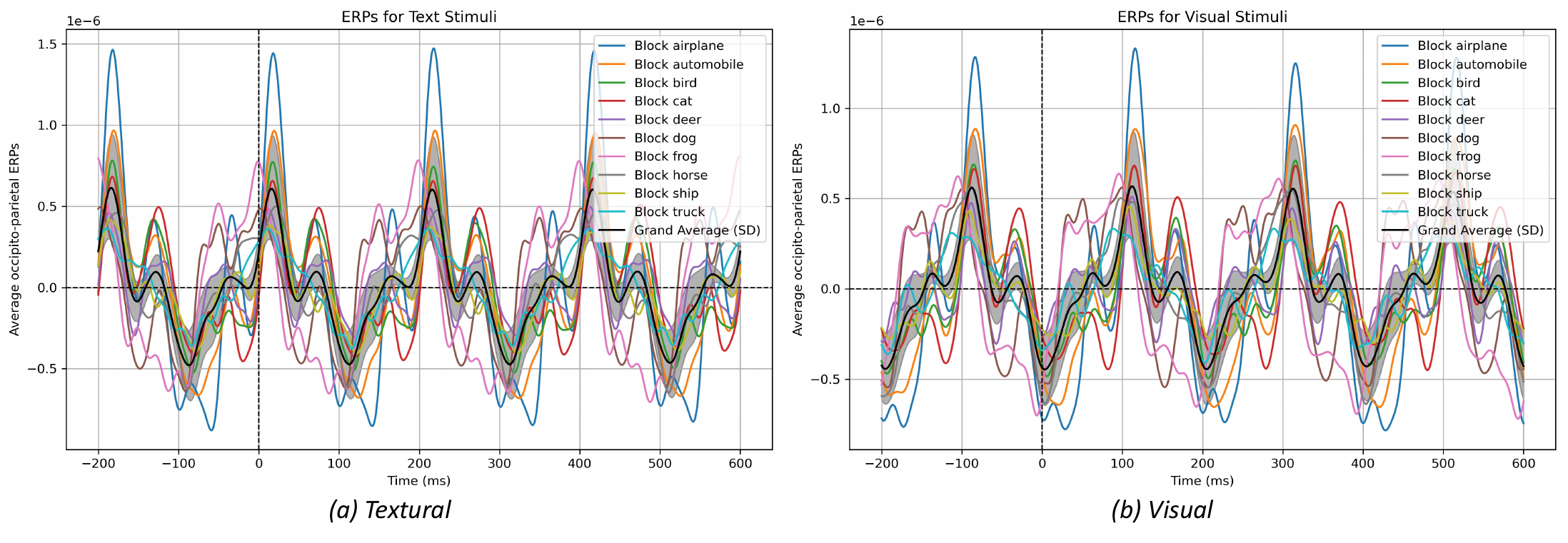}
    %\vspace{-24pt}
    \caption{ERPs averaged over occipital and parietal electrodes for the participant viewing stimuli from (a) visual images and (b) the category text. Shaded areas around the grand average ERP represent standard deviations at each time point.}
    \label{fig:erp}
    %\vspace{-16pt}
\end{figure*}
% \noindent \textbf{Discussion}
Unlike previous EEG datasets, such as the THINGS-EEG dataset~\cite{THINGSEEG1}, which use high-resolution images as visual stimuli and introduce a vast number of object concepts (1854), our datasets can address the following limitations of previous ones. Fig.~\ref{fig:resolution} illustrates the differences in EEG signal responses between high-resolution and lower-resolution visual stimuli. The more stable and less variable neural responses to the lower-resolution images suggest their suitability for creating robust EEG datasets. High-resolution images, on the other hand, require more time for participants to process content and details, making them less suitable for effectively capturing quick neural responses at the millisecond level.

\subsection{ERP Analysis}

Fig.~\ref{fig:erp} presents the event-related potentials (ERPs) averaged over occipital and parietal electrodes for a participant viewing visual images (right panel) and category text (left panel). Both plots display ERP data from -200 ms to 600 ms relative to stimulus onset (0 ms), with the average occipito-parietal ERPs fluctuating between approximately -0.5 and 1.5 microvolts for both visual and text stimuli. Each trace, with a distinct color, represents a specific category, including airplane, automobile, bird, cat, deer, dog, frog, horse, ship, and truck.

Regarding the text stimuli (left panel), a significant initial deflection is noticed around 0 ms, showing the brain's quick response to text stimuli. Early components, like peaks and troughs, are seen around 100 ms and 200 ms post-stimulus, typical of early ERP components such as the P1 and N1, which are linked to sensory processing. Additional peaks around 300 ms and beyond likely indicate higher-order cognitive processing. The shaded area around the grand average ERP line signifies the standard deviation, reflecting variability across different trials and categories. This variability is higher at certain peaks, suggesting differences in how the brain processes various text categories.

Concerning the visual stimuli (right panel), a comparable initial deflection is observed around 0 ms. Distinct peaks are evident at approximately 100 ms and 200 ms, corresponding to the P1 and N1 components, which are more pronounced and consistent across different visual categories compared to textual stimuli. Significant peaks around 300 ms and later may denote the P3 component, indicating cognitive processing associated with visual categorization. The standard deviation shading around the grand average indicates less variability compared to text stimuli, suggesting more consistent brain responses to visual stimuli across various categories.

In comparison, visual stimuli evoke more consistent ERPs across categories than text stimuli, as indicated by the smaller standard deviation areas. Both types of stimuli elicit similar amplitude ranges in the ERP responses, reflecting comparable levels of neural activity. The timing of early and late ERP components is similar for both text and visual stimuli, suggesting that initial sensory processing and subsequent cognitive processing occur within similar time frames for both types of stimuli.
In conclusion, the ERPs for both textual and visual stimuli exhibit characteristic early and late components, indicative of sensory and cognitive processing stages. Visual stimuli elicit more consistent responses across categories, whereas text stimuli exhibit greater variability. This analysis provides insights into the sensory and cognitive functions associated with different types of stimuli.
\begin{table*}[t]
    \centering 
    %\vspace{-5pt}
    \renewcommand{\tabcolsep}{12pt}
    \resizebox{\linewidth}{!}{
    \begin{tabular}{l|ccc|ccc|ccc}
    \toprule
        \multirow{2}*{Models} & \multicolumn{3}{c|}{Image} & \multicolumn{3}{c|}{Text} & \multicolumn{3}{c}{Image \& Text} \\
        \cmidrule{2-10}
        &Acc & Recall & F1 &Acc & Recall & F1 &Acc & Recall & F1 \\
         \midrule
         EEGNet~\cite{lawhern2018eegnet} & 25.42 & 25.63 & 24.75 & 25.62 & 26.30 & 24.77 & 20.72 & 20.96 & 19.69\\
        MobileNet\_v2~\cite{sandler2018mobilenetv2}& 40.84 & 41.61 & 40.79& 40.17 & 39.64 & 39.52 & 49.76 & 49.57 & 49.32\\
    ResNet18~\cite{he2016deep}& \textbf{56.57} & \textbf{56.41} & \textbf{56.46} & 56.38 & 56.17 & 56.17 & \textbf{63.53} & \textbf{63.65} & \textbf{63.55}\\
    ResNet34~\cite{he2016deep}& 56.41 & 56.15 & 56.24 & \textbf{56.47} & \textbf{56.34} & \textbf{56.39} & 58.77 & 59.22 & 58.89\\
    ResNet50~\cite{he2016deep}&49.45 & 48.49 & 48.80 & 49.93 & 49.46 & 49.61 & 49.34 & 50.26 & 49.38\\
    \bottomrule

    \end{tabular}
    }
    \caption{Benchmark experiments within one session of one participant.}
    \label{tab:one_participant}
    %\vspace{-10pt}
\end{table*}

\begin{table*}[t]
    \centering
    %\vspace{-5pt}
    \renewcommand{\tabcolsep}{12pt}
    \resizebox{\linewidth}{!}{
    \begin{tabular}{l|ccc|ccc|ccc}
    \toprule
        \multirow{2}*{Models} & \multicolumn{3}{c|}{Image} & \multicolumn{3}{c|}{Text} & \multicolumn{3}{c}{Image \& Text} \\
        \cmidrule{2-10}
        &Acc & Recall & F1 &Acc & Recall & F1 &Acc & Recall & F1 \\
         \midrule
         EEGNet~\cite{lawhern2018eegnet}&22.08 & 22.24 & 21.80&24.15 & 24.11 & 23.56&20.73 & 20.68 & 19.83 \\
        MobileNet\_v2~\cite{sandler2018mobilenetv2}& 41.90 & 42.18 & 41.66 &42.67 & 43.32 & 41.59& 48.97 & 49.28 & 48.86\\
        ResNet18~\cite{he2016deep}& 53.49 & \textbf{53.73} & \textbf{53.45}& \textbf{54.11} & \textbf{54.10} & \textbf{54.10} & 58.60 & 58.65 & 58.54\\ 
        ResNet34~\cite{he2016deep}& \textbf{54.06} & 53.27 & 52.42& 53.57 & 53.30 & 52.65& \textbf{60.69} & \textbf{60.91} & \textbf{60.76}\\
        ResNet50~\cite{he2016deep}& 49.80 & 49.60 & 48.53& 49.82 & 48.81 & 48.17& 56.07 & 54.98 & 54.98\\

    \bottomrule
    \end{tabular}
    }
    \caption{Benchmark experiments across different sessions of two participants.}
    %\vspace{-16pt}
    \label{tab:two_participant}
\end{table*}
\begin{figure*}[t!]
    \centering
    \includegraphics[width=\textwidth]{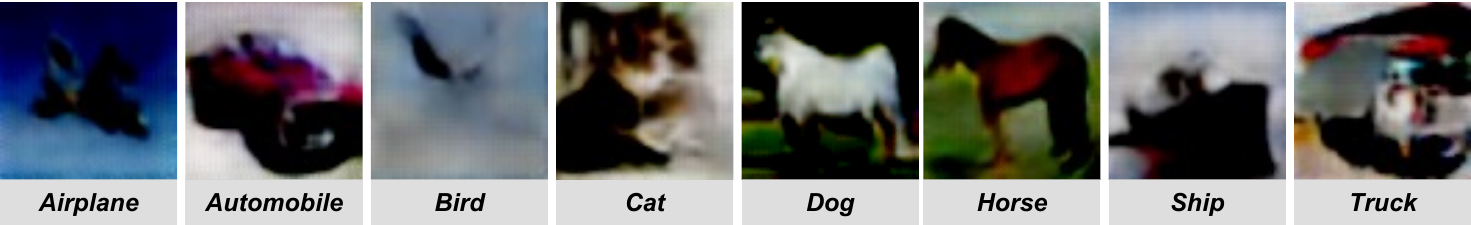}
    %\vspace{-20pt}
    \caption{Generation results of our dataset using the ThoughtVis Model.}
    %\vspace{-12pt}
    \label{fig:gen}
\end{figure*}
%\vspace{-5pt}
\section{Experiments with EIT-1M Dataset}
%\vspace{-10pt}
% To assess the effectiveness of the proposed EIT-1M, we conduct experiments on both recognition and generation tasks. 
\label{Experiments}
\noindent \textbf{Implementation Details.}
For preprocessing, a band-pass filter is applied to retain frequencies between 1 and 40 Hz within the raw EEG data. Subsequently, the continuous data is segmented into epochs, each commencing 50 ms prior to the stimulus onset and concluding 50 ms following each event. To train and evaluate the recognition models, the EEG data from one participant (Tab.~\ref{tab:one_participant}) and two participants (Tab.~\ref{tab:two_participant})  are divided using an 80/20\% split to create training and evaluation sets, respectively. The models are trained using the Adam optimizer, coupled with a step learning rate schedule, across 500 epochs. The default settings for the learning rate, weight decay, and batch size are \(1 \times 10^{-3}\), \(1 \times 10^{-5}\), and 2048, respectively. We apply three widely-used metric to evaluate the recognition performance on EIT-1M, including Accuracy, recall, and F1 score.
% Code can be found in \url{https://eit-1m.github.io/EIT-1M/}. 
% \noindent \textbf{Evaluation Metrics.}
% \textbf{Accuracy} (Acc) measures the proportion of correct predictions out of the total number of predictions.
% % It is useful when classes are balanced but can be misleading with imbalanced datasets, as it doesn't account for class distribution.
% \textbf{Recall}, also known as sensitivity, measures the proportion of actual positive instances that are accurately identified. It is crucial to minimize false negatives.
% % , especially in applications like medical diagnostics where missing a positive case is critical.
% The \textbf{F1} score is the harmonic mean of Acc and recall, effectively balancing the trade-off between the two metrics.%  It is useful for imbalanced datasets, considering both false positives and false negatives, and provides a more balanced evaluation of model performance.

%\vspace{-5pt}
\subsection{Recognition}
%\vspace{-5pt}
\label{sec:recogntion}

The results of experiments conducted within one session of a single participant are shown in Tab.~\ref{tab:one_participant}, illustrating the effectiveness of our dataset in the individual collection procedure. The results in Tab.~\ref{tab:one_participant} include the performance across various models with EEG signals captured from visual and textual stimuli. Note that $Image \& Text$ refers to the combined EEG signals from both visual and textual stimuli for recognition. The evaluated models include EEGNet~\cite{lawhern2018eegnet}, MobileNet-v2~\cite{sandler2018mobilenetv2}, ResNet18~\cite{he2016deep}, ResNet34~\cite{he2016deep}, and ResNet50~\cite{he2016deep}.

Combining EEG signals from image and text stimuli generally enhances performance metrics across all models, suggesting that multi-modal data provides richer information, leading to better classification accuracy and robustness. 
The consistent performance improvements observed from MobileNet-v2 to ResNet architectures indicate that our EIT-1M dataset is well-suited for various deep-learning models. ResNet models, in particular, show significant improvements, highlighting the dataset's capacity to support complex neural networks. Similar performance metrics for image and text stimuli alone indicate that the dataset offers a balanced representation of both modalities. This balance is crucial for training models to generalize well across different types of stimuli. Additionally, the high F1 scores, especially for the ResNet models, reflect good data quality, ensuring that the recorded EEG signals are reliable and effective for training AI models.
Tab.~\ref{tab:two_participant} summarizes benchmark experiments across different sessions of two participants. The results consistently show that combining EEG signals from both visual and textual stimuli improves performance across all models compared to using either visual or textual stimuli alone. For both visual and textual stimuli, ResNet models maintain consistently high performance, indicating the robustness of ResNet architectures in processing and learning from EEG data.

The analysis of Tab.~\ref{tab:one_participant} and Tab.~\ref{tab:two_participant} supports the rationality of our EIT-1M dataset. By providing high-quality, balanced, and scalable data, our dataset proves to be an excellent resource for advancing research in multi-modal AI and cognitive neuroscience. The observed improvements in combined image and text stimuli further highlight the importance of multi-modal datasets in capturing the intricate interplay between different types of information.

% \begin{figure}
%     \centering
%     \includegraphics[width=1\linewidth]{images/benchmark_classification.pdf}
%     \caption{Enter Caption}
%     \label{fig:enter-label}
% \end{figure}

%\vspace{-10pt}
\subsection{Generation}
\label{sec:generation}
% We follow the classic EEG-to-Image generation work ThoughtVis~\cite{tirupattur2018thoughtviz} to obtain EEG-generated images. Due to the space limit, please refer to the supplementary materials. \textcolor{red}{Analysis of Fig. 6}
We follow the classic EEG-to-Image generation task proposed by ThoughtVis~\cite{tirupattur2018thoughtviz}, which obtains images from EEG signals. As shown in the generation results in Fig.~\ref{fig:gen}, our proposed EIT-1M dataset shows the capability to support the EEG-to-Image generation task.

%\vspace{-10pt}
\section{Conclusion, Limitations, and Future Work}
\label{Limitations}
In this paper, we presented EIT-1M, a large-scale multi-modal dataset comprising 1 million EEG-image-text pairs. We collected the data pairs while participants viewed alternating sequences of visual-textual stimuli from 60K natural images and corresponding label texts. Our EIT-1M is superior in its capacity of recording brain activities in simultaneously processing multi-modal information, \ie, images and text. It features response-based stimulus timing and repetition across blocks and sessions. To verify the effectiveness of EIT-1M, we provided an in-depth analysis of the EEG signals in EIT-1M across different categories and sessions and conducted experiments on two tasks.

\noindent \textbf{Limitations.} Despite the robustness of our dataset, there are areas for enhancement. Our current dataset includes data from multiple participants and sessions, but increasing the number of participants and sessions could yield a more comprehensive understanding of neural responses and improve the generalizability of the models trained on this data. Additionally, while we used a well-defined set of visual and textual stimuli, expanding the variety of stimuli, especially for the textual stimuli, could further enhance the dataset's fidelity for studying more diverse and complex neural processes.

\noindent \textbf{Future work.} It could be a good direction to integrate additional modalities, such as audio or tactile feedback, to create an even richer multi-modal dataset. This integration could provide deeper insights into the interplay between different sensory inputs and brain activity, advancing research in multi-modal AI and neuroscience. By addressing these limitations and expanding the dataset's scope, we can significantly contribute to the understanding and development of multi-modal AI models.

\noindent \textbf{Broader Impact.} EIT-1M advances neuroscience and AI by enabling deeper insights into cognitive processes and sensory integration. It improves brain-computer interfaces and personalized learning. Ethical considerations regarding neural data privacy are crucial for responsible applications.

% \clearpage
{
    \small
    \bibliographystyle{ieeenat_fullname}
    \bibliography{main}
}

\end{document}